\documentclass[runningheads]{llncs}
\usepackage[T1]{fontenc}
\usepackage{graphicx}
\begin{document}

\title{BioRAGent: A Retrieval-Augmented Generation System for Showcasing Generative Query Expansion and Domain-Specific Search for Scientific Q\&A}
\titlerunning{BioRAGent: domain-specific RAG for Q\&A}
\author{Samy Ateia\inst{1}\orcidID{0009-0000-2622-9194}\email{}\and
Udo Kruschwitz\inst{1}\orcidID{0000-0002-5503-0341}}

\authorrunning{S. Ateia and U. Kruschwitz}
\institute{University of Regensburg, Regensburg, Germany\\
\email{\{Samy.Ateia,Udo.Kruschwitz\}@sprachlit.uni-regensburg.de}\\
}
\maketitle              % typeset the header of the contribution

\begingroup
\renewcommand\thefootnote{}
\footnotetext{
Version as accepted at the Demo Track at ECIR 2025 \\
2024 Copyright for this paper by its authors. 
Use permitted under Creative Commons License Attribution 4.0 International (CC BY 4.0).
}
\addtocounter{footnote}{-1}
\endgroup

\begin{abstract}
We present BioRAGent, an interactive web-based retrieval-augmented generation (RAG) system for biomedical question answering. The system uses large language models (LLMs) for query expansion, snippet extraction, and answer generation while maintaining transparency through citation links to the source documents and displaying generated queries for further editing. Building on our successful participation in the BioASQ 2024 challenge, we demonstrate how few-shot learning with LLMs can be effectively applied for a professional search setting. The system supports both direct short paragraph style responses and responses with inline citations. Our demo is available online\footnote{\url{https://bioragent.samyateia.de/}}, and the source code is publicly accessible through GitHub\footnote{\url{https://github.com/SamyAteia/BioRAGent}}.

\end{abstract}
\section{Introduction}
Professional search in biomedical domains requires complex query formulation to support the generation of evidence-based answers \cite{MACFARLANE2022200091}. While LLMs have shown impressive capabilities in question answering tasks, their tendency to hallucinate makes it challenging to directly use them in professional settings \cite{ji2023survey}. RAG \cite{lewis2020retrieval} has emerged as a promising approach to ground LLM responses in reliable sources and reduce hallucinations \cite{shuster2021retrieval}.

Building on our participation in the BioASQ 2024 challenge \cite{BioASQ2024overview}, where we demonstrated competitive performance using both commercial and open-source LLMs in a biomedical RAG setting \cite{AteiaK24}, we present \textbf{BioRAGent} - an interactive system that makes our approach accessible to users through a web interface. The system combines generative query expansion, snippet extraction, and answer generation while maintaining transparency through direct links to PubMed source documents.

This interactive demonstration system serves three \textbf{purposes}: showcasing in-context learning capabilities in specialized domains, demonstrating RAG implementations in professional search, and evaluating LLM performance in biomedical query processing. The \textbf{scope} encompasses biomedical question-answering tasks, with a focus on evidence-based responses grounded in PubMed literature. The primary \textbf{audience} comprises academic researchers studying LLM applications in domain-specific search and biomedical search professionals evaluating AI-powered solutions. Through transparent query expansion and citation processes, users can assess the utility of current LLMs for complex biomedical information needs \cite{10.1145/3490238}.

\section{System Overview}

BioRAGent is implemented as a web application using the Gradio framework\footnote{\url{https://www.gradio.app/}}, which provides a clean and responsive user interface. The underlying LLM used is Gemini 1.5 flash 002 from Google due to its speed and low cost\footnote{\url{https://developers.googleblog.com/en/updated-gemini-models-reduced-15-pro-pricing-increased-rate-limits-and-more/}} \cite{reid2024gemini}. The system architecture consists of four main components:

\subsection{Query Expansion}
Given a biomedical question, the system uses few-shot (3-shot) learning with LLMs to generate an expanded query that incorporates relevant synonyms and related terms. The potential of LLMs generating queries for professional search has been explored in related work, notably by Wang et al. (2023) \cite{wang2023can}. Users can inspect and modify the expanded query after execution, making the search process transparent and controllable. The system uses Elasticsearch as the underlying search engine with an index of the 2023 snapshot of PubMed articles \footnote{\url{https://pubmed.ncbi.nlm.nih.gov/download/\#annual-baseline}}. The query syntax is the Elasticsearch \emph{query string query dsl}\footnote{\url{https://www.elastic.co/guide/en/elasticsearch/reference/current/query-dsl-query-string-query.html\#query-string-syntax}} that is supported by the used search endpoint.

\subsection{Document Retrieval and Snippet Extraction}
The expanded query is executed against an index of abstracts and titles of PubMed articles using the default English analyzer of Elasticsearch\footnote{\url{https://www.elastic.co/guide/en/elasticsearch/reference/current/analysis-lang-analyzer.html\#english-analyzer}}. The top 50 articles ranked by the default BM25 based scoring of Elasticsearch are retrieved and for each document, the LLM is prompted with few-shot examples to extract relevant snippets. The system processes multiple documents in parallel to maintain acceptable performance. In the last step, the LLM is again prompted with few-shot examples to rerank the extracted snippets based on their relevance to the original question. Few-shot examples in this step are sourced from the BioASQ training set and not optimized, except for query expansion, where they are sampled by highest f1 score. 

\subsection{Answer Generation}
The system generates two types of answers:
\begin{itemize}
    \item A short paragraph style answer that is generated by grounding the model with the retrieved snippets but doesn't force the model to use the retrieved information. This format was used in the BioASQ challenge.
    \item A paragraph style answer with inline citations to PubMed articles from the retrieved snippets in the form of PubMed IDs for every generated sentence. For this answer format the model needs to use the retrieved snippets and cannot generate an answer if no relevant snippets were found. This format was used in the TREC 2024 BioGen Track\footnote{\url{https://dmice.ohsu.edu/trec-biogen/}}.
\end{itemize}

\subsection{User Interface}
The interface provides:
\begin{itemize}
    \item A question input field with a search button
    \item An editable text box to display the Expanded query
    \item Two answer text boxes with and without citations
    \item A list of retrieved snippets with direct PubMed links
\end{itemize}

\begin{figure}[h]
    \centering
    \includegraphics[width=\textwidth]{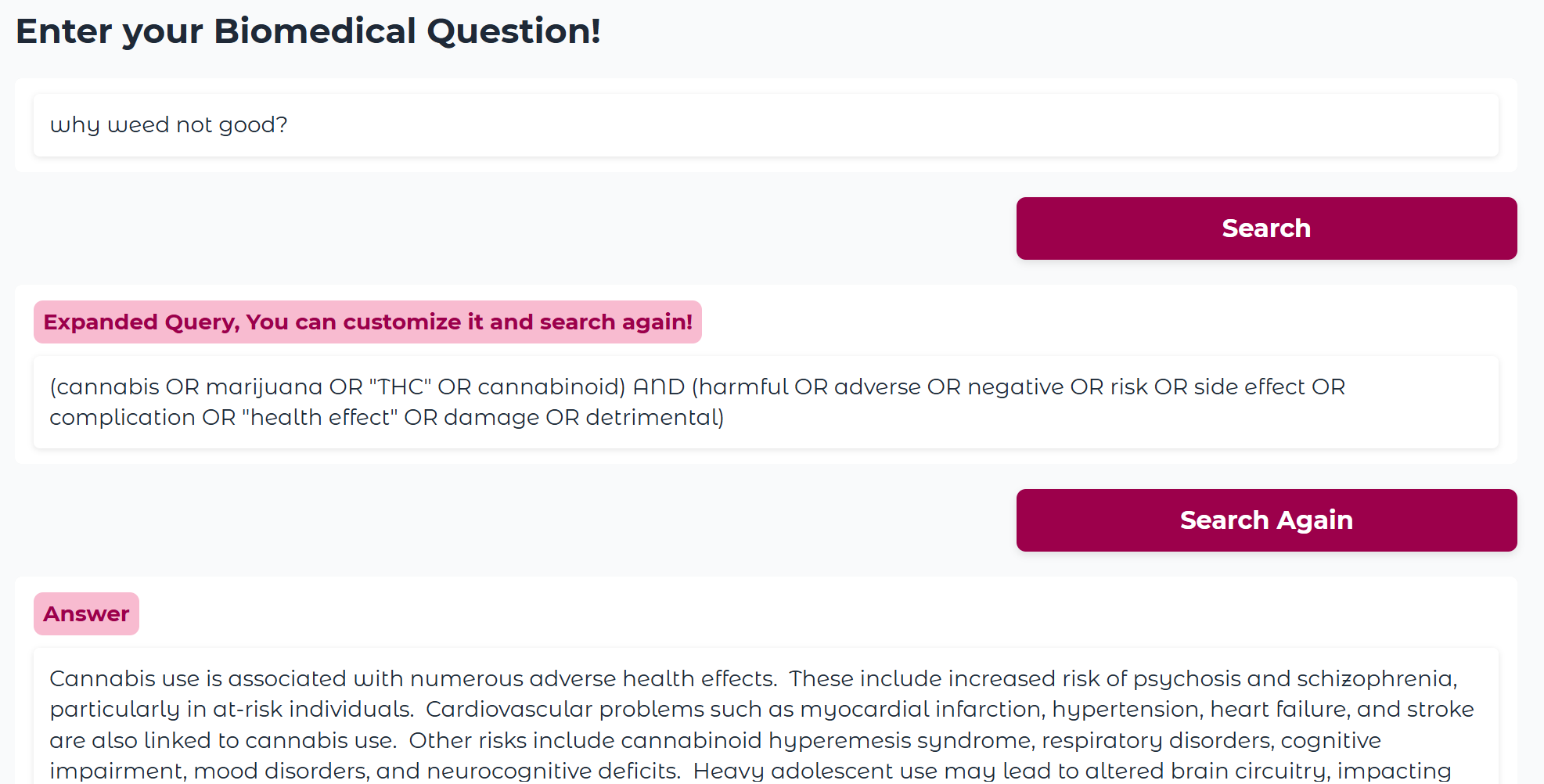} 
    \caption{Screenshot of part of the BioRAGent interface, showcasing query expansion}
    \label{fig:interface}
\end{figure}

A screenshot highlighting the ability of the system to turn overly simple questions into query terms that actually fit the target domain (scientific papers) is showcased in Figure \ref{fig:interface}.

\section{Evaluation and Discussion}
The core RAG components of BioRAGent were evaluated through our participation in the BioASQ 2024 challenge \cite{AteiaK24}. Our system achieved competitive results across different question formats and settings, winning multiple first and second places in the 12th. BioASQ challenge\footnote{\url{https://bioasq.org/participate/twelfth-challenge-winners}}.

While our system was most competitive in the question answering tasks of the challenge (12B Phase A+, Phase B), in the document retrieval and snippets extraction tasks other systems that also used dense and hybrid retrieval techniques took the leading spots.

While dense vector search is an intriguing technique due to the demonstrated state-of-the-art (SOTA) performance in multiple benchmarks, the resulting retrieval rankings based on embedding vectors are not transparent or easily controllable by a search expert. Our approach, on the other hand, makes the encoded semantic knowledge of the LLM used for searching, visible and controllable by displaying the expanded query.

\section{Conclusion and Future Work}
BioRAGent demonstrates how SOTA LLM capabilities could be integrated into professional search systems while maintaining transparency and source attribution. The system makes our successful BioASQ approach accessible to users through an intuitive interface.

We hope that showcasing the usefulness of our rather simple in-context learning approaches, will encourage others to build upon and rapidly refine these methods, advancing the state-of-the-art in scientific and biomedical RAG. Future work will focus on expanding the interface for live prompt and few-shot template editing; incorporating direct evaluation modules for BioASQ dataset verification and hallucination detection; and adding support for selecting and evaluating different LLMs.

\begin{credits}
\subsubsection{\ackname} 
We thank the BioASQ and TREC BioGen organizers for enabling expert evaluation of our approaches through their competitions.
\subsubsection{\discintname}
The authors have no competing interests to declare that are relevant to the content of this article. 
\end{credits}
\bibliographystyle{splncs04}
\bibliography{bibliography}

\begin{thebibliography}{1}
\providecommand{\url}[1]{\texttt{#1}}
\providecommand{\urlprefix}{URL }
\providecommand{\doi}[1]{https://doi.org/#1}

\bibitem{AteiaK24}
Ateia, S., Kruschwitz, U.: {Can Open-Source LLMs Compete with Commercial Models? Exploring the Few-Shot Performance of Current {GPT} Models in Biomedical Tasks}. In: Faggioli, G., Ferro, N., Galusc{\'{a}}kov{\'{a}}, P., de~Herrera, A.G.S. (eds.) Working Notes of the Conference and Labs of the Evaluation Forum {(CLEF} 2024), Grenoble, France, 9-12 September, 2024. {CEUR} Workshop Proceedings, vol.~3740, pp. 78--98. CEUR-WS.org (2024), \url{https://ceur-ws.org/Vol-3740/paper-07.pdf}

\bibitem{ji2023survey}
Ji, Z., Lee, N., Frieske, R., Yu, T., Su, D., Xu, Y., Ishii, E., Bang, Y.J., Madotto, A., Fung, P.: {Survey of Hallucination in Natural Language Generation}. ACM Comput. Surv.  \textbf{55}(12) (2023). \doi{10.1145/3571730}

\bibitem{10.1145/3490238}
Jin, Q., Yuan, Z., Xiong, G., Yu, Q., Ying, H., Tan, C., Chen, M., Huang, S., Liu, X., Yu, S.: {Biomedical Question Answering: A Survey of Approaches and Challenges}. ACM Comput. Surv.  \textbf{55}(2) (Jan 2022). \doi{10.1145/3490238}, \url{https://doi.org/10.1145/3490238}

\bibitem{lewis2020retrieval}
Lewis, P., Perez, E., Piktus, A., Petroni, F., Karpukhin, V., Goyal, N., K{\"u}ttler, H., Lewis, M., Yih, W.t., Rockt{\"a}schel, T., et~al.: Retrieval-augmented generation for knowledge-intensive nlp tasks. Advances in Neural Information Processing Systems  \textbf{33},  9459--9474 (2020)

\bibitem{MACFARLANE2022200091}
MacFarlane, A., Russell-Rose, T., Shokraneh, F.: Search strategy formulation for systematic reviews: Issues, challenges and opportunities. Intelligent Systems with Applications  \textbf{15},  200091 (2022). \doi{https://doi.org/10.1016/j.iswa.2022.200091}

\bibitem{BioASQ2024overview}
Nentidis, A., Katsimpras, G., Krithara, A., Lima-López, S., Farré-Maduell, E., Krallinger, M., Loukachevitch, N., Davydova, V., Tutubalina, E., Paliouras, G.: {Overview of BioASQ 2024: The twelfth BioASQ challenge on Large-Scale Biomedical Semantic Indexing and Question Answering}. In: Goeuriot, L., Mulhem, P., Quénot, G., Schwab, D., Soulier, L., Maria Di~Nunzio, G., Galuščáková, P., García Seco~de Herrera, A., Faggioli, G., Ferro, N. (eds.) Experimental IR Meets Multilinguality, Multimodality, and Interaction. Proceedings of the Fifteenth International Conference of the CLEF Association (CLEF 2024) (2024)

\bibitem{reid2024gemini}
Reid, M., Savinov, N., Teplyashin, D., Lepikhin, D., Lillicrap, T., Alayrac, J.b., Soricut, R., Lazaridou, A., Firat, O., Schrittwieser, J., et~al.: {Gemini 1.5: Unlocking multimodal understanding across millions of tokens of context}. arXiv preprint arXiv:2403.05530  (2024)

\bibitem{shuster2021retrieval}
Shuster, K., Poff, S., Chen, M., Kiela, D., Weston, J.: {Retrieval Augmentation Reduces Hallucination in Conversation}. In: Findings of the Association for Computational Linguistics: EMNLP 2021. pp. 3784--3803 (2021)

\bibitem{wang2023can}
Wang, S., Scells, H., Koopman, B., Zuccon, G.: {Can ChatGPT write a good boolean query for systematic review literature search?} In: Proceedings of the 46th International ACM SIGIR Conference on Research and Development in Information Retrieval. pp. 1426--1436 (2023)

\end{thebibliography}
%\end{thebibliography}
\end{document}